\documentclass[conference]{IEEEtran}
\usepackage{authblk}
\IEEEoverridecommandlockouts
\usepackage{url}
\usepackage{cite}
\usepackage{amsmath,amssymb,amsfonts}
\usepackage{algorithmic}
\usepackage{graphicx}
\usepackage{textcomp}
\usepackage{xcolor}
\usepackage{multirow}
\usepackage{algorithmic}
\usepackage{mathrsfs}
\usepackage{booktabs}
\usepackage{color}
\usepackage{bm}
\usepackage{float}
\usepackage{pifont}

\def\BibTeX{{\rm B\kern-.05em{\sc i\kern-.025em b}\kern-.08em
    T\kern-.1667em\lower.7ex\hbox{E}\kern-.125emX}}

\begin{document}

\title{\huge A Byte Sequence is Worth an Image: CNN for File Fragment Classification Using Bit Shift and n-Gram Embeddings}

\author[1]{Wenyang Liu}
\author[1]{Yi Wang}
\author[1]{Kejun Wu}
\author[1]{Kim-Hui Yap}
\author[2]{Lap-Pui Chau\vspace{-0.8em}} 
\affil[1]{\textit {School of Electrical and Electronics Engineering, Nanyang Technological University, Singapore}}
\affil[2]{\textit {Dept. of Electronic and Information Engineering, The Hong Kong Polytechnic University, Hong Kong}}
\affil[ ]{\tt \small  {\{wenyang001, wang1241\}@e.ntu.edu.sg, \{kejun.wu, ekhyap\}@ntu.edu.sg, lap-pui.chau@polyu.edu.hk}\vspace{-1.5em}}

\renewcommand\Authands{ and }
\maketitle

\begin{abstract}
File fragment classification (FFC) on small chunks of memory is essential in memory forensics and Internet security. Existing methods mainly treat file fragments as 1d byte signals and utilize the captured inter-byte features for classification, while the bit information within bytes, i.e., intra-byte information, is seldom considered. This is inherently inapt for classifying variable-length coding files whose symbols are represented as the variable number of bits. Conversely, we propose Byte2Image, a novel data augmentation technique, to introduce the neglected intra-byte information into file fragments and re-treat them as 2d gray-scale images, which allows us to capture both inter-byte and intra-byte correlations simultaneously through powerful convolutional neural networks (CNNs). Specifically, to convert file fragments to 2d images, we employ a sliding byte window to expose the neglected intra-byte information and stack their n-gram features row by row. We further propose a byte sequence \& image fusion network as a classifier, which can jointly model the raw 1d byte sequence and the converted 2d image to perform FFC. Experiments on FFT-75 dataset validate that our proposed method can achieve notable accuracy improvements over state-of-the-art methods in nearly all scenarios. The code will be released at \url{https://github.com/wenyang001/Byte2Image}.

\end{abstract}


\section{Introduction}
\label{sec:intro}
File carving is an important memory forensic technique, aiming to recover meaningful files from storage devices without the file system metadata. Due to the file system fragmentation~\cite{pal2009evolution}, file carving methods need to scan the whole storage devices to find possible file fragments and reassemble them into complete files. To reduce the search space, file fragment classification (FFC) methods~\cite{amirani2008new,mittal2020fifty} are proposed, which can identify the file type directly based on the raw data of file fragments so that only file fragments of the same file type need to be considered during each file reassembly. Meanwhile, FFC also plays a crucial role in Internet security applications, such as anti-viruses, anti-spam, intrusion detection systems, and firewalls~\cite{amirani2008new, khodadadi2021classification}. A good FFC is essential.




A simple and straightforward idea is to rely on magic bytes~\cite{pal2009evolution} (e.g, from file headers or footers) to identify the file type, but this method suffers from seriously fragmented files that do not contain magic bytes. Hand-crafted features~\cite{beebe2013sceadan} show more reliable performance since they consider inherent statistics of different file types, such as byte frequency analysis and byte frequency cross-correlation~\cite{mittal2020fifty}. However, it is complicated for humans to design different hand-crafted features for various file types. 





\begin{figure}[t]
\centering
\includegraphics[width=3.4in]{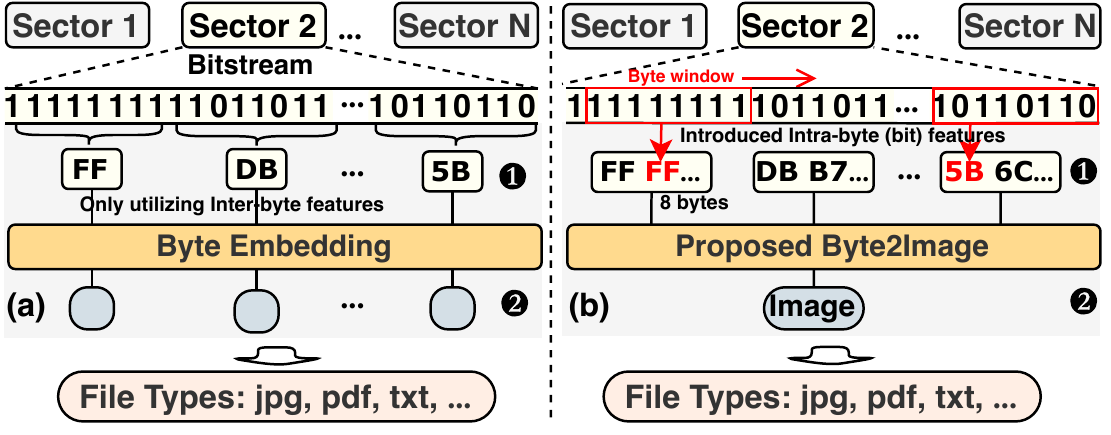}  
\caption{\footnotesize Comparison on existing file fragment classification (FFC) methods with our proposed FFC method. (a) Existing methods~\cite{mittal2020fifty,haque2022byte} only consider the inter-byte features. (b) Two insights of ours, i.e., the intra-byte features are introduced by using a sliding byte window (stride=1 bit), and the file fragment is re-treated as a 2d image to be processed by powerful CNNs.}
\label{f:insights}
\vspace{-0.15in}
\end{figure}

\begin{figure}[t]
\centering
\includegraphics[width=3.4in]{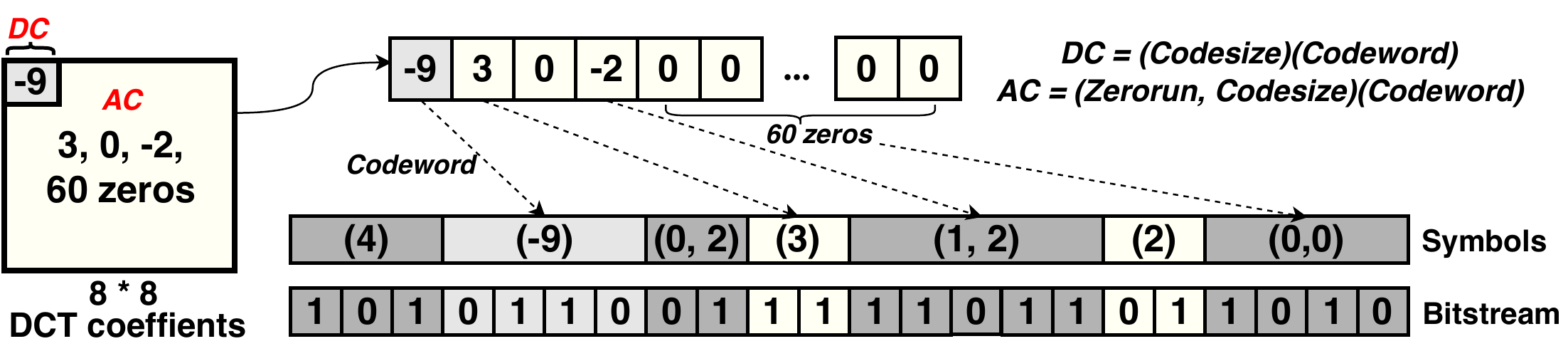}  
\caption{\footnotesize Illustration of Huffman coding steps used in JPEG files, where intermediate symbols of DCT coefficients are represented as variable-length bits.} 
\label{f:jpeg}
\vspace{-0.25in}
\end{figure}

Recently, with the advance in machine learning, data-driven techniques have shown great promise in FFC that enables implicit feature modeling auto-learned by the neural network. Wang \textit{et al.}~\cite{wang2018sparse} used sparse coding to enable automated feature extraction and a support vector machine (SVM) for classification. Motivated by the word embedding in natural language processing (NLP), a novel feature embedding model called Byte2Vec~\cite{haque2022byte} was proposed, which used the Skip-Gram model~\cite{mikolov2013distributed} to learn the dense vector representation for each byte and adopted the k-Nearest Neighbor (kNN) for classification. However, since the feature extraction and the classification are separated, the overall performance is limited by the individual optimization of each module. In contrast to Byte2Vec, Mittal~\textit{et al.} proposed an end-to-end method called FiFTy~\cite{mittal2020fifty} that can jointly optimize feature extraction and classification by using 1d CNN, building a new baseline for FFC. Saaim~\textit{et al.}~\cite{saaim2022light} further proposed a lightweight depthwise separable convolution (DSCNN) for FFC.

Despite performance improvements made by data-driven methods, there is one critical issue that has been severely neglected, i.e., the bit information within bytes. Existing works simply treat memory sectors of file fragments as 1d byte signals (see bytes of FF and D8 in Fig.~\ref{f:insights}(a) are generated by grouping every 8 bits, respectively), and only consider the inter-byte relations even if learning a suitable byte embedding. This is inapt for classifying files that employ variable-length codes (VLCs) to compress data, e.g., Huffman codes in JPEG files where encoded symbols are not packed into integer bytes as Fig.~\ref{f:jpeg} shows.

In view of the above issues, we propose Byte2Image to introduce the neglected intra-byte information and re-treat file fragments as 2d gray-scale images. Specifically, we first employ a byte sliding window (stride = 1 bit) shown in \ding{182} of Fig.~\ref{f:insights}(b) to expose intra-byte information as additional bytes. Each byte in Fig.~\ref{f:insights}(a) now corresponds to a byte sequence of eight bytes in Fig.~\ref{f:insights}(b). Next, we stack these byte sequences row by row to construct a 2d image, which allows us to capture both inter-byte and intra-byte correlations by powerful CNNs. However, we found CNN's performance suffers from the low aspect ratio (width to height) of the constructed image since most commonly used CNNs are designed for square images~\cite{ghosh2019reshaping}. Inspired by n-grams used in NLP, we introduce intra-byte n-grams in Byte2Image to make the converted image in \ding{183} of Fig.~\ref{f:insights}(b) more square, i.e., higher aspect ratio. Finally, we propose a byte sequence \& image fusion network as a classifier that enables a joint modeling of the raw 1d byte sequence and the converted 2d image to perform FFC. Our classifier is instantiated as a simple fully connected (FC) layer and a powerful CNN backbone. The FC layer is to consider the magical byte co-occurrence from the raw 1d byte sequence and the CNN backbone is to capture the complex inter-byte and intra-byte correlations from the converted image. 


\begin{figure*}[t]
\centering
\includegraphics[width=7.1in]{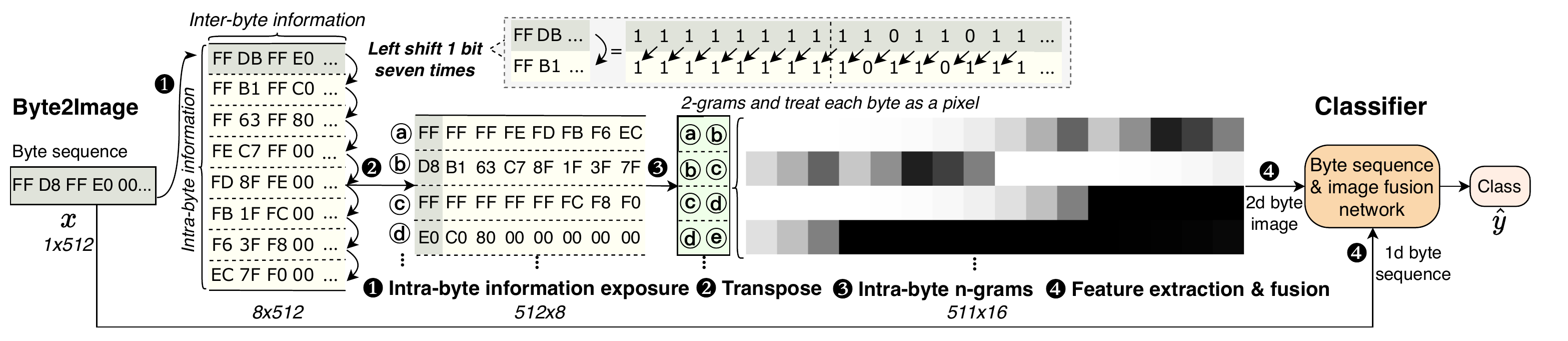} 
\caption{\footnotesize Architecture of our proposed method. In Byte2Image, the sliding byte window is implemented by bit-shifting the raw byte sequence for seven times, respectively, to uncover the intra-byte information. The intra-byte information is stacked row by row to convert the file fragment into a 2d byte matrix. We then use intra-byte n-grams to increase the aspect ratio of the byte matrix and view it as a gray-scale image by regarding each byte as a pixel (\textit{FF} corresponds to white color and \textit{00} corresponds to black color). Note transpose operation here is for better illustration. In classifier, our proposed network can consider both the 1d byte sequence and the converted 2d gray-scale image modalities to achieve file fragment classification.}
\label{f:BER_b}
\vspace{-0.15in}
\end{figure*}


Our contributions are summarized below. (1) To the best of our knowledge, this is the first work considering bit information within bytes, i.e., intra-byte information, in file fragment classification. (2) We first propose Byte2Image to introduce the neglected intra-byte information into file fragments and re-treat them as 2d gray-scale images. And the proposed byte sequence \& image fusion network enables a joint modeling of the raw 1d byte sequence and the converted 2d image to perform file fragment classification. (3) Extensive experiments on the published FFT-75 dataset~\cite{mittal2020fifty} validate that our method can achieve state-of-the-art results.

\section{The proposed method}
Given a byte sequence $x$ of the memory sector with $N_s$ bytes, where $x\in \mathbb{Z}_{255}^{N_s}$, $\mathbb{Z}_{255} = \{0, 1, .., 255\}$, and $N_s$ is 512 or 4,096 based on the file systems, the target is to estimate the corresponding file-type label $y$. We address this problem by introducing the intra-byte information into file fragments and re-treating them as 2d gray-scale images. Considering that the converted 2d images provide a new visual modality, we further propose a byte sequence \& image fusion network as a classifier to consider the joint byte sequence and converted image modalities. Fig.~\ref{f:BER_b} shows the overall architecture of our method consisting of two main parts.

\subsection{Byte2Image}



The overview of our proposed Byte2Image architecture is shown in Fig.~\ref{f:BER_b}, which contains two modules.

\subsubsection{Sliding byte window} The employed sliding byte window, i.e., window = 8 bit, stride = 1 bit, is implemented by bit-shifting the raw byte sequence $x$ for seven times, respectively (see \ding{182} in Fig.~\ref{f:BER_b}), to uncover the intra-byte information. We then convert $x$ to a 2d byte matrix $\mathbf{x_{in}}$ by stacking intra-byte information row by row and transposing the byte matrix for better illustration (see \ding{183} in Fig.~\ref{f:BER_b}), expressed as $x \rightarrow{} \mathbf{x_{in}} \in \mathbb{Z}_{255}^{N_s \times 8}$, where $\mathbf{x_{in}} = [x_0^\mathsf{T}, x_1^\mathsf{T}, ..., x_7^\mathsf{T}]$, $x_0 = x \in \mathbb{Z}_{255}^{N_s}$ and $x_i = x_{i-1} << 1$. For the byte matrix $\mathbf{x_{in}}$, each row consists of eight bytes and adjacent bytes differ by 1-bit distance, representing the neglected \textbf{\textit{intra-byte information}}, and each column consists of $N_s$ bytes where adjacent bytes differ by 8-bit distance, i.e., 1-byte distance, representing the \textbf{\textit{inter-byte information}}. Hence, both intra- and inter-byte information is included in $\mathbf{x_{in}}$.

\subsubsection{Intra-byte n-grams} For the converted byte matrix $\mathbf{x_{in}}$, an intuitive idea is to treat it as a gray-scale image directly as a pixel of an 8-bit image exactly occupies one byte, and extract features through CNNs. Therefore, $\mathbf{x_{in}}$ can be viewed as a gray-scale image with a $N_s \times 8$ resolution. However, since most commonly used CNNs~\cite{ghosh2019reshaping}, e.g., resnet~\cite{he2016deep} and densenet~\cite{huang2017densely}, are designed for square images, too small of the aspect ratio (the ratio of width to height) $8 / N_s$ of this gray-scale image makes it harder for these CNNs to extract the complex inter-byte and intra-byte correlations. Inspired by the n-gram features~\cite{kim-2014-convolutional,tripathy2016classification} used in NLP for modeling the sentence of text, we regard each row of $\mathbf{x_{in}}$, i.e., intra-byte information, as a unigram feature and use its n-grams to increase the image width. This is to say a new gray-scale image $\mathbf{x_{ngram}}$ is constructed by stacking these intra-byte n-grams row by row (see \ding{184} in Fig.~\ref{f:BER_b} where \textcircled{a} and \textcircled{b} are concatenated as 2-grams in $\mathbf{x_{ngram}}$ for example), expressed as $\mathbf{x_{in}} \rightarrow{} \mathbf{x_{ngram}} \in \mathbb{Z}_{255}^{H \times W \times 1}$, where $H = N_s - n + 1$ and $W=8n$. There are two obvious benefits of using intra-byte n-grams. First, the aspect ratio of $\mathbf{x_{ngram}}$ become higher than $\mathbf{x_{in}}$, which is better for CNNs to extract features. Second, adjacent bytes in intra-byte n-grams still differ by 1-bit distance and hence contain more relative order information, which is better for classification.

\subsection{Classifier} 

As the 1d raw byte sequence and the converted 2d image can be viewed as two different modalities or types of information, it is a good idea to develop a model that can take both into account. Similar to~\cite{cheng2016wide, zheng2017wide}, we adopt a two-stream design and propose a new byte sequence \& image fusion network.







\begin{figure}[t]
\centering
\includegraphics[width=3.4in]{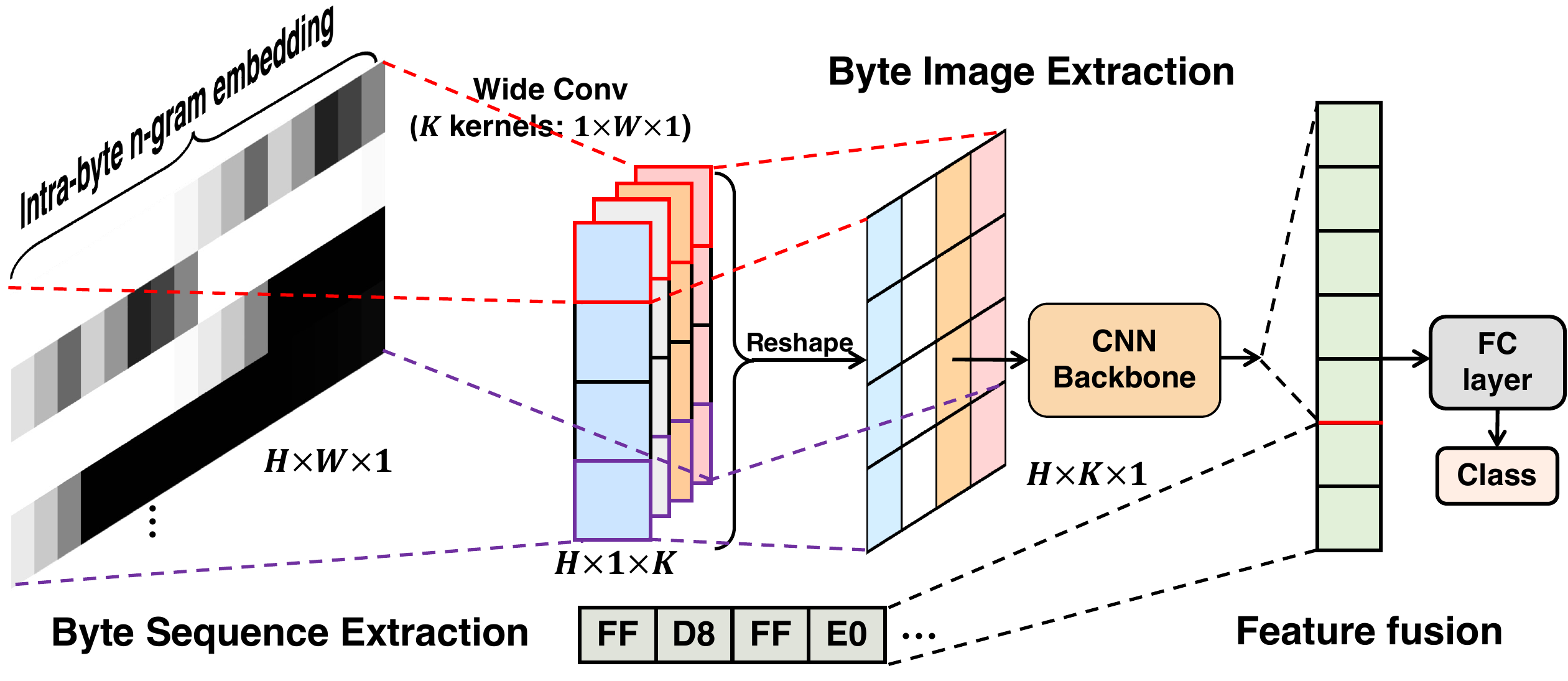}  
\caption{\footnotesize Byte sequence \& image fusion network.} 
\label{f:down}
\vspace{-0.15in}
\end{figure}



\subsubsection{Byte sequence extraction} As shown in Fig.~\ref{f:down}, we use one FC layer to process the 1d raw byte sequence. As we mentioned, magic bytes of file fragments can be a useful indicator for file fragment classification, which can be loosely defined as the frequent co-occurrence of some special bytes. It turns out that a FC layer can memorize global features well in~\cite{cheng2016wide, zheng2017wide}. In this case, we expect that the employed one FC layer can directly memorize the most relevant co-occurrence of magic bytes for classification, expressed as:


\vspace{-0.1in}
\begin{equation}
\mathbf{x_{byte}} = \text{ReLU}(\mathbf{w} x + b)
\end{equation} 

\noindent where $x \in \mathbb{Z}_{255}^{N_s}$ is the initial byte sequence, $\mathbf{w}$ represents trainable parameters of the FC layer and $b$ is the bias.

\subsubsection{Byte image extraction} As shown in Fig.~\ref{f:down}, the module input is the converted 2d gray-scale image with intra-byte n-grams. Normally, it will be followed by n-gram embeddings for each n-gram~\cite{kim-2014-convolutional,tripathy2016classification} to better fuse the n-gram information and make the model to have more flexibility. As a result, similar to~\cite{kalchbrenner2014convolutional}, we introduce a wide convolution (kernel width is equal to the image width) to embed each row of sparse \textit{intra-byte n-grams} to a dense row vector by channel concatenate. Specifically, a wide convolution operation involves a learned weight matrix $\mathbf{w_i} \in \mathbb{R}^{1 \times W \times 1}$ over the image $\mathbf{x_{ngram}} \in \mathbb{Z}_{255}^{H \times W \times 1}$ to produce a 1d embedding feature $\mathbf{x_{emb, i}}$. A total of $K$ weight matrices are applied to get $K$ embeddings. These embeddings are then concatenated to get a 2d feature map $\mathbf{x_{emb}} \in \mathbb{R}^{H \times K \times 1}$, expressed as:

\vspace{-0.25in}
\begin{gather}
    \mathbf{x_{emb, i}} = \mathbf{w_i} \mathbf{x_{ngram}} + b_i  \\
    \mathbf{x_{emb}} = [\mathbf{x_{emb, 1}}, \mathbf{x_{emb, 2}}, ..., \mathbf{x_{emb, K}}]
\end{gather}

\vspace{-0.05in}
\noindent For the new 2d feature map $\mathbf{x_{emb}}$, we then apply a commonly used powerful deep CNN backbone, e.g., resnet18~\cite{he2016deep}, to extract the most relevant inter-byte and intra-byte correlations for classification. Feature maps of the last layer of the CNN backbone can be expressed as: 

\vspace{-0.15in}
\begin{equation}
\mathbf{x_{cnn}} = \text{CNN}(\mathbf{x_{emb}})
\end{equation}

\subsubsection{Feature fusion} The outputs of different modalities are concatenated to generate the final features by using a weighted sum, and then fed to \textit{Softmax} layer to get predicted probabilities for each class, expressed as:

\vspace{-0.15in}
\begin{equation}
\hat{y} = \textit{Softmax}(\mathbf{w} [\mathbf{x_{byte}}, \mathbf{x_{cnn}}] + b)
\end{equation}

\noindent where $\mathbf{w}$ represents joint trainable parameters of the last FC layer, $b$ is the bias.

\subsection{Model for 4k-byte sector}
An issue with our model processing the 4k-byte sector is that it has a high computational complexity since the converted image $\mathbf{x_{ngram}}$ is up to 4k resolution with the height $H=N_s-n+1$. A naive solution is to use a downsampling module or resize operation to reduce the computation burden. However, we find that both operations lead to much information reduction and cause severe performance degradation. Here, observing that the 512-byte sector can already be a good classifier, we divide the 4k-byte memory sector into eight 512-byte parts and convert each part into a gray-scale image in the same way by using the Byte2Image technique. The eight gray-scale images are then concatenated over the channel and sent into CNN backbones so that the high computational button can be decreased, while the other modules keep the same for the file fragment classification.




\section{Experiments}
\subsection{Experimental settings}
\subsubsection{Dataset} We conduct our experiments on the FFT-75 dataset which is the largest corpus for file fragment classification to date. It contains 75 file types, and each type has 102,400 samples for 512- and 4,096-byte sector sizes. The dataset consists of 6 scenes for different use cases: \#1 for all 75 file-type classifications, \#2 for 11 group classifications, \#3 for photos \& videos carver with 25 classes, \#4 for coarse photo carver with \#5 classes (i.e., JPEG, RAW photos, videos, bitmaps, and others), \#5 for specialized JPEG carver with 2 classes (i.e., JPEG and other 74 file types), and \#6 for camera specialized JPEG carver with 2 classes (i.e., JPEG and other RAW photos \& videos). 

\begin{table*}[h]    
\centering

\caption{\footnotesize Comparison with FIFTy~\cite{mittal2020fifty} and DSCNN~\cite{saaim2022light} in a sector size (SS) of 512 and 4,096 bytes from scenario \#1 to \#6 of FFT-75, where the number of classes of each scenario required to be classified is given. “-” means there is no JPEG class in this scenario}

\label{T:scen2}
\resizebox{\linewidth}{!}{
\begin{tabular}{lccccccccccccc} 
\hline
\multirow{2}{*}{Method}  & \multirow{2}{*}{SS} &    \multicolumn{2}{c}{75 classes (\#1)} &\multicolumn{2}{c}{11 classes (\#2)}   & \multicolumn{2}{c}{25 classes (\#3)}    & \multicolumn{2}{c}{5 classes (\#4)}  &  \multicolumn{2}{c}{2 classes (\#5)}   & \multicolumn{2}{c}{2 classes (\#6)}  \\ \cline {3-14}
        &    &  Acc. & JPEG Acc & Acc. & JPEG Acc. & Acc. & JPEG Acc. & Acc. & JPEG Acc. & Acc. & JPEG Acc. & Acc. & JPEG Acc.  \\ \hline  

FiFTy~\cite{mittal2020fifty}    & 512     & 65.6 & 83.5 & 78.9 & - &  87.9 & 93.3 & 90.2 & \textbf{98.6} &  99.0 & 99.3  & \textbf{99.3} & \textbf{99.5}  \\ 

DSCNN~\cite{saaim2022light} & 512 & 65.9 & 83.4 &  74.8 &  - &  80.8  & 94.6 & 87.2 & 97.3 & 98.9 & 98.9 & 98.8 & 98.6 \\

Ours (ResNet18) &   512  & \textbf{71.0} & \textbf{89.2} &  \textbf{90.4} & - &  \textbf{93.5} & \textbf{96.6} & \textbf{93.6} & 97.4 & \textbf{99.2} & \textbf{99.3}  & 99.2 & 99.2 \\ \hline   
FiFTy~\cite{mittal2020fifty}    & 4,096 & 77.5 & 86.3 &  89.8  & - & 94.6 & \textbf{98.9} &  94.1 & \textbf{99.1} & 99.2 & 99.2 & \textbf{99.6} & \textbf{99.7}   \\ 

DSCNN~\cite{saaim2022light} & 4,096 & 78.5 & 91.7 & 85.7 & - & 93.1 & 96.8 & 94.2 & 98.4 & 99.3 & 99.3 & 99.6 & 99.7 \\

Ours (ResNet18) & 4,096 & \textbf{82.1}  & \textbf{93.4}  & \textbf{94.2}   & - & \textbf{96.8} & 98.1 & \textbf{96.1}  & 98.6 & \textbf{99.3} & \textbf{99.6} & 99.4  & 99.5   \\ \hline
\end{tabular}
}
\vspace{-0.1in}
\end{table*}



\subsubsection{Implementation Details} The sector size $N_s$ is set to 512 or 4,096 based on the dataset. The n-gram $n$ is set to 16 and the dimension for n-gram embeddings $K$ is set to 96. For CNN backbones, we choose resnet-18/-34~\cite{he2016deep} to test performance. For the image pre-processing, we choose the Random Horizontal Flip and Random Erasing~\cite{zhong2020random} as augmentation techniques. As for training, we adopt the AdamW~\cite{loshchilov2017decoupled} optimizer with $(\beta_1, \beta_2)=(0.9, 0.999)$ and weight decay equal to 0.01. The batch size is set to 1024. The learning rate is warmed up from 0 to $5\text{e-}4$ linearly in the first 2 epochs and then decays to 0 via the cosine scheduler in the rest 48 epochs. We implement our model with the PyTorch on two NVIDIA GeForce RTX 3090 GPUs to train our model. 




\subsection{Experimental results}
\subsubsection{Comparison on FFT-75 of All Scenarios} Following~\cite{mittal2020fifty}, we report the classification accuracy for all file types on testing datasets. Table~\ref{T:scen2} shows the comparison of the results with the state-of-the-art FiFTy~\cite{mittal2020fifty} and DSCNN~\cite{saaim2022light}. All models deliver better results at SS = 4,096 than SS = 512 since the large sector size contains more information. Our model outperforms FiFTy and DSCNN a lot, especially in complicated scenarios \#1 and \#2, which demonstrates the importance of the introduced neglected intra-byte information. Compared with the FiFTy baseline, our method improves the average accuracy by 5.4\% and 11.5\% in SS = 512 of scenarios \#1 and \#2, achieving the best results with the accuracy of 71.0\% and 90.4\%, respectively. As the number of classes required to be classified decreases, the problem becomes much easier. All methods can perform well in file-type limited scenarios, i.e., \#4 to \#6, and can achieve high accuracy of nearly 99\% over JPEG accuracy in 2-class classification (JPEG-vs-others), i.e., \#5 and \#6 scenarios. In addition, our model adopts the same architecture for all scenarios, but FiFTy and DSCNN optimize the architecture (e.g., layer number and kernel size) according to different scenarios and sector sizes. Therefore, our model is more adaptable and flexible than FiFTy.


\subsubsection{Ablation Studies} 
To further validate our model, more state-of-the-art methods are evaluated on the most complicated scenario of FFT-75, i.e., scenario \#1 for SS = 512. The comparing approaches include three methods trained with hand-crafted features (i.e., Sceadan, NN-CF and NN-CO) and three methods trained with byte-embedded features from the 1d raw byte sequence only (i.e., Byte2Vec+kNN, FiFTy and DSCNN). As Table~\ref{T:scen1} shows, since our method considers both features from the raw 1d byte sequence and the converted 2d gray-scale image with the neglected intra-byte information, we obtain the best results than other state-of-the-art methods. We also ablate our model with different CNN backbones. The results show that the employed resnet-34 achieves better than the employed resnet-18, achieving the best results with the accuracy of 71.4\% and 91.8\%, respectively. It demonstrates that our model can benefit from the powerful CNNs, which proves our idea of converting byte sequences to images. It is worth noting that our model may have extra accuracy improvements by replacing with more powerful CNNs, showing our model's flexibility.

\begin{table}[]
\centering
\caption{\footnotesize Comparison with more state-of-the-art methods in sector size (SS) of 512 bytes in scenario \#1 of FFT-75. Sceadan, NN-GF and NN-CO are using hand-crafted features. Byte2Vec, FiFTy and DSCNN are using byte-embedded features of 1d raw byte sequence only}

\label{T:scen1}
\footnotesize
\begin{tabular}{l|ccc|cc} \hline
 \multirow{2}{*}{Method}  & \multicolumn{3}{c|}{Features}  & \multicolumn{2}{c}{SS = 512 (\#1)} \\ \cline{2-6} 
&  Hand & 1d   &  2d  & Acc.  & JPEG Acc. \\ \hline
Sceadan~\cite{beebe2013sceadan} &   \checkmark  & &  & 57.3  & 81.5  \\ 
NN-GF~\cite{mittal2020fifty} &   \checkmark  & & & 45.4  & 53.2  \\ 
NN-CO~\cite{mittal2020fifty} &    \checkmark  &  & & 64.4  & 77.0 \\ \hline 
Byte2Vec+kNN~\cite{haque2022byte} & & \checkmark  &  & 50.1 & 42.4\\  
FiFTy~\cite{mittal2020fifty} & & \checkmark   & & 65.6 & 83.5  \\
DSCNN~\cite{saaim2022light} & & \checkmark   & & 65.9 & 83.4 \\ \hline
Ours (ResNet18)  & & \checkmark   & \checkmark &\textbf{71.0} & \textbf{89.2}  \\  
Ours (ResNet34)  & & \checkmark   & \checkmark &\textbf{71.4} & \textbf{91.8}  \\ \hline
\end{tabular}
\vspace{-0.1in}
\end{table}

\section{Conclusion}
This paper proposes Byte2Image, a novel data augmentation technique, to introduce the neglected intra-byte information into file fragments and re-treat them as 2d gray-scale images. As the 1d raw byte sequence and the converted 2d image can be seen as two different modalities, we further propose a new byte sequence \& image fusion network to take both modalities into account, where the simple magic bytes co-occurrence can be captured from 1d raw byte sequence and the complex inter-byte and intra-byte correlations can be captured by powerful CNNs. Extensive experiments show our method exceeds state-of-the-art methods on file type classification, which demonstrates the superiority of our method.

\section*{Acknowledgement}
This research/project is supported by the National Research Foundation, Singapore, and Cyber Security Agency of Singapore under its National Cybersecurity R\&D Programme (NRF2018NCR-NCR009-0001). Any opinions, findings and conclusions or recommendations expressed in this material are those of the author(s) and do not reflect the views of National Research Foundation, Singapore and Cyber Security Agency of Singapore.




\bibliographystyle{IEEEtran}
\bibliography{refs}
\end{document}